\documentclass[11pt,a4paper]{article}

\usepackage{times}
\usepackage[T1]{fontenc}
\usepackage[utf8]{inputenc}
\usepackage{geometry}
\usepackage{graphicx}
\usepackage{booktabs}
\usepackage{multirow}
\usepackage{natbib}
\usepackage{hyperref}
\usepackage{url}
\usepackage{xcolor}
\usepackage{tabularx}
\usepackage{array}
\usepackage{enumitem}
\usepackage{titlesec}
\usepackage{fancyhdr}
\usepackage{microtype}
\usepackage{float}
\usepackage{placeins}

\hypersetup{
 colorlinks=true,
 linkcolor={blue!80!black},
 citecolor={blue!80!black},
 urlcolor={blue!80!black}
}

\twocolumn

\titleformat{\section}{\normalsize\bfseries}{\thesection}{1em}{}
\titleformat{\subsection}{\normalsize\bfseries}{\thesubsection}{1em}{}
\titleformat{\subsubsection}{\normalsize\bfseries\itshape}{\thesubsubsection}{1em}{}
\titlespacing*{\section}{0pt}{12pt plus 2pt minus 1pt}{6pt plus 1pt}
\titlespacing*{\subsection}{0pt}{10pt plus 2pt minus 1pt}{4pt plus 1pt}
\titlespacing*{\subsubsection}{0pt}{8pt plus 2pt minus 1pt}{4pt plus 1pt}

\usepackage[font=small,labelfont=bf]{caption}

\begin{document}

\twocolumn[
\begin{center}
{\Large\bfseries Oral to Web: Digitizing `Zero Resource'\\Languages of Bangladesh\par}
\vspace{12pt}
{\large\bfseries Mohammad Mamun Or Rashid\par}
\vspace{4pt}
{\normalsize Consultant, EBLICT Project, Bangladesh Computer Council\\
Associate Professor, Jahangirnagar University\\
\texttt{mamunbd@juniv.edu}, \texttt{mamunbd@bcc.gov.bd}\par}
\vspace{12pt}

\begin{minipage}{0.92\textwidth}
\begin{center}
{\normalsize\bfseries Abstract}
\end{center}
\small
This paper presents the Multilingual Cloud Corpus, the first national-scale, parallel, multimodal linguistic dataset of Bangladesh's ethnic and indigenous languages. The corpus comprises 85,792 structured textual entries---each containing a Bengali stimulus text, its English translation, and an IPA transcription---together with approximately 107 hours of transcribed audio recordings, spanning 42 language varieties from four families (Tibeto-Burman, Indo-European, Austro-Asiatic, Dravidian) and two unclassified languages. Constructed through systematic fieldwork across nine districts involving 77 speakers and 43 validators, the dataset covers isolated lexical items, grammatical constructions, and directed speech across 97 thematic categories. We describe the methodology of converting these predominantly oral languages into structured, web-accessible digital resources on the Multilingual Cloud platform (\url{https://multiling.cloud}), discuss patterns in endangered language documentation, and assess the corpus's implications for computational linguistics and NLP in extremely low-resource settings.
\end{minipage}
\vspace{16pt}
\end{center}
]

\section{Introduction}

Bangladesh is widely perceived as a linguistically homogeneous nation. Bengali, the constitutionally recognized national language and a focal point of the country's independence movement, is spoken as a first language by approximately 98\% of its 170 million citizens \citep{eberhard2024}. Beneath this apparent uniformity, however, lies a stratum of linguistic diversity that remains poorly documented. Of the approximately 40 ethnic minority languages spoken in Bangladesh, 14 have been identified as endangered by the International Mother Language Institute \citep{imli2018}. Yet prior documentation efforts have been predominantly descriptive and community-specific, focusing on individual languages or small language groups (e.g., \citealp{akter2024}). No systematic, parallel corpus has been constructed across all four language families at a national scale. The absence of such a resource has impeded both computational linguistics research and community-driven revitalization for these languages.

The title of this work, ``Oral to Web,'' encapsulates the central trajectory of the project: transforming languages that exist predominantly in oral form, with little or no written tradition or digital footprint, into structured, web-accessible digital resources. This process entails converting spoken utterances captured through systematic fieldwork into annotated textual entries with phonetic transcriptions, parallel translations, and segmented audio, all hosted on a publicly accessible web platform. In doing so, the project bridges the gap between the oral transmission practices through which these languages have historically survived and the digital infrastructure required for their computational processing, scholarly analysis, and long-term preservation.

This paper addresses this gap by presenting the \textbf{Multilingual Cloud Corpus}, the first national-scale, parallel, multimodal linguistic dataset of Bangladesh's ethnic and indigenous languages. The corpus comprises 85,792 structured textual entries, each containing a Bengali stimulus text, its English translation, and an IPA transcription, together with approximately 107 hours of transcribed audio recordings, spanning 42 language varieties from four families and two unclassified languages. The corpus makes several contributions to the field. It is the first attempt to systematically digitize ethnic languages of Bangladesh across all four represented language families at a national scale.

\section{Literature Review}

This section situates our work within three intersecting strands of prior scholarship: (i) the ethnolinguistic taxonomy and classification of minority languages in Bangladesh, (ii) the broader enterprise of endangered language documentation and corpus construction, and (iii) the computational resource landscape for low-resource and extremely low-resource languages in South Asia.

\subsection{Ethnolinguistic Diversity and Language Taxonomy in Bangladesh}

Bangladesh harbours considerable linguistic diversity among its ethnic minority communities. The 2022 national census enumerated 1.65 million people from 50 ethnic minority groups, constituting less than 1.1\% of the total population \citep{bbs2022}. These communities are concentrated in four principal geographic zones: the Chittagong Hill Tracts in the southeast, the Sylhet Division in the northeast, the Rajshahi Division in the west, and the Mymensingh Division in north-central Bangladesh. The earliest systematic documentation of the languages spoken in what is now Bangladesh dates to Grierson's \textit{Linguistic Survey of India} (1903--1928), which catalogued Tibeto-Burman, Munda, Dravidian, and Indo-Aryan varieties across the Indian subcontinent. Volume 3 of the survey provided the first comparative vocabulary lists and grammatical sketches for several Tibeto-Burman languages of the Chittagong Hill Tracts, including varieties ancestral to modern Marma, Chakma, and Kuki-Chin languages \citep{grierson1903}. However, no comparably extensive linguistic survey of the region has been conducted since, as noted by \citet{khan2010} and others who have observed that Bangladesh's constitutional and policy frameworks have historically failed to acknowledge the country's linguistic diversity.

\begin{figure*}[!htb]
\centering
\includegraphics[width=\textwidth,height=0.55\textheight,keepaspectratio]{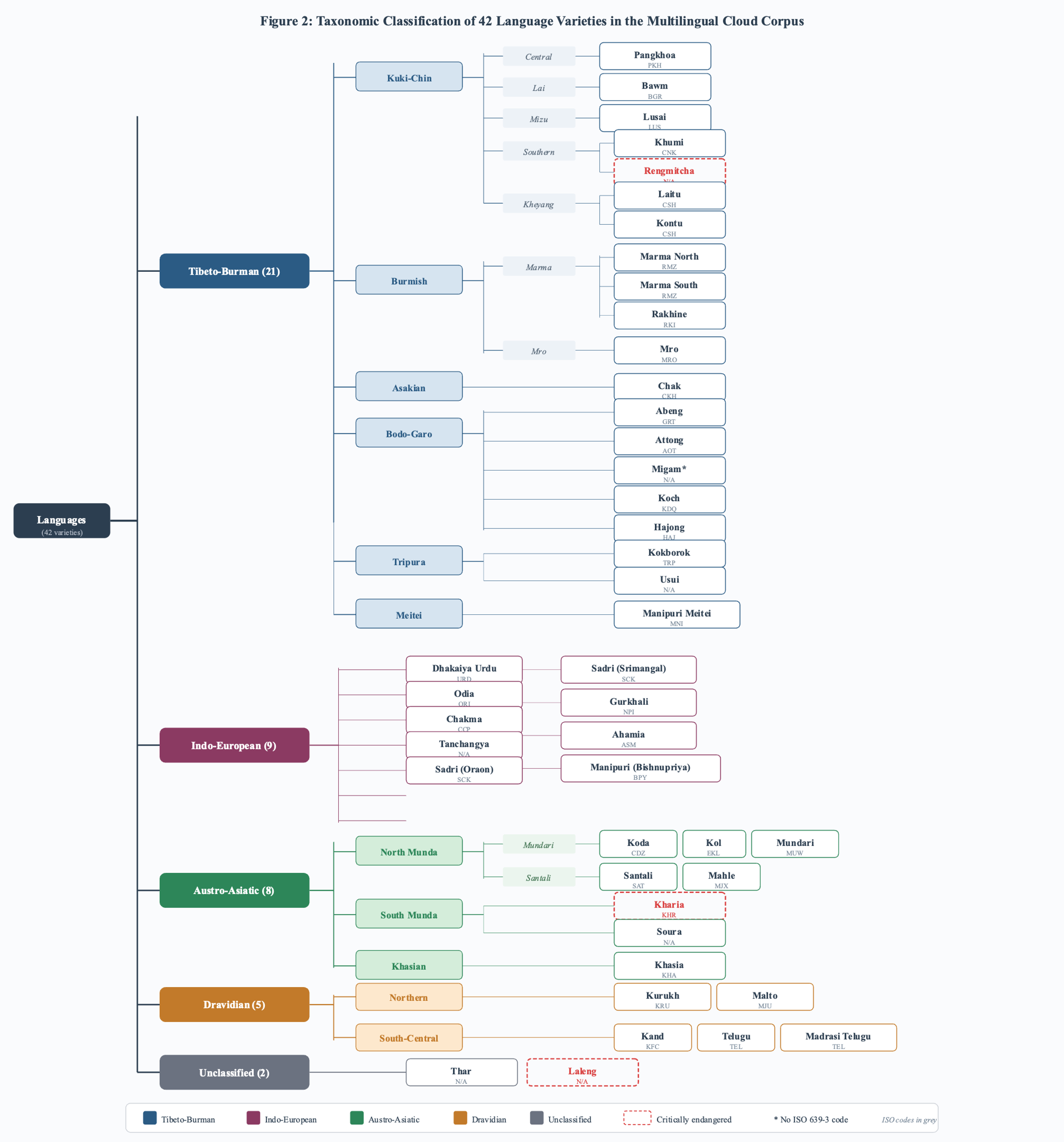}
\caption{Taxonomic Classification of Language Varieties}
\label{fig:taxonomy}
\end{figure*}

The Tibeto-Burman family constitutes the largest group with 21 language varieties in our corpus, organized into six branches. The Kuki-Chin branch is the most diverse, further subdivided into five sub-branches: Central (Pangkhoa), Lai (Bawm), Mizu (Lusai), Southern (Khumi, Rengmitcha), and Kheyang (Laitu, Kontu). Rengmitcha, classified as a Khomic language of the Southern sub-branch \citep{peterson2017}, is critically endangered with approximately six remaining speakers. The Burmish branch comprises the Marma sub-branch (Marma North, Marma South, Rakhine) and the Mro sub-branch (Mro). The Asakian branch is represented by Chak, an endangered language with its own indigenous script. The Bodo-Garo branch includes the three Garo dialect varieties (Abeng, Attong, and Migam), as well as Koch and Hajong. The Tripura branch comprises Kokborok and Usui, closely related varieties of the Tripura language complex. The Meitei branch is represented by Manipuri Meitei, which employs the Meitei Mayek indigenous script. \citet{faquire2020} has documented extensive language contact phenomena in the Chittagong Hill Tracts, where Tibeto-Burman languages such as Marma, Chak, Khumi, and Kheyang have undergone significant structural convergence with Indo-Aryan varieties due to prolonged contact with Bengali-speaking populations. \citet{akter2024} provides an overview of Pangkhoa as a South Central Tibeto-Burman (Kuki-Chin) language, contributing to the descriptive documentation of this branch.

The Indo-European family is represented by 9 varieties in our corpus: Dhakaiya Urdu, Odia, Chakma, Tanchangya, Sadri (Oraon), Sadri (Srimangal), Gurkhali, Ahamia, and Manipuri (Bishnupriya). Chakma and Tanchangya, while spoken in the Chittagong Hill Tracts alongside Tibeto-Burman languages, are linguistically closer to Chittagonian Bengali rather than to the Tibeto-Burman languages of the region. Bishnupriya Manipuri is spoken in the Sylhet Division. Sadri is represented by two distinct community varieties such as Sadri (Oraon) spoken in the Rajshahi Division and Sadri (Srimangal) associated with tea garden worker communities in Sylhet. Dhakaiya Urdu represents the historic Urdu-speaking community of Old Dhaka.

The Austro-Asiatic family is represented by 8 varieties across three branches. The North Munda branch includes the Mundari sub-group (Koda, Kol, Mundari) and the Santali sub-group (Santali, Mahle). The South Munda branch includes Kharia and Soura. The Khasian branch is represented by Khasia, spoken in the Sylhet borderlands and constituting the northernmost Austroasiatic language. The Munda branch languages are spoken primarily by communities historically associated with tea garden labour in the Sylhet Division and by Adivasi communities in the northern districts.

The Dravidian family is represented by 5 varieties across two branches. The Northern branch includes Kurukh (Oraon) and Malto (Sauria Paharia), spoken by communities in the Rajshahi and Rangpur divisions of western Bangladesh, with Kurukh speakers also present in Sylhet tea garden communities. Both languages face significant pressure from language shift toward Bengali, particularly among younger generations \citep{eberhard2024}. The South-Central branch includes Kand, Telugu, and Madrasi Telugu. Telugu and Madrasi (Telugu) are maintained as separate corpus entries to preserve sociolinguistic distinctions between the Telugu community in Moulvibazar and the Madrasi community in Dhaka.

Two languages, Thar and Laleng, remain unclassified in our taxonomy due to ongoing scholarly debate about their genetic affiliation. Laleng (also known as Patro) is critically endangered.

\subsection{Language Documentation and Corpus Construction for Endangered Languages}

The theoretical foundations of modern language documentation were articulated by \citet{himmelmann1998,himmelmann2006}, who distinguished language documentation as the creation of multipurpose, lasting records of a language from language description and the production of grammars and dictionaries. Himmelmann's framework emphasizes the creation of audio-visual corpora that capture language use across a range of communicative contexts, genres, and speakers, producing records that can serve both the research community and the speech communities themselves. This paradigm has been further elaborated by \citet{austin2011}, who survey the methodological, ethical, and technological dimensions of documenting endangered languages, and by \citet{woodbury2011}, who articulates principles of corpus adequacy including diversity, balance, and naturalness. A persistent challenge in endangered language documentation is the tension between standardized, cross-linguistically comparable elicitation and the documentation of naturalistic language use. Our work addresses this by incorporating both structured elicitation (words and sentences) and semi-naturalistic directed speech across thematically diverse conversational scenarios, an approach that draws on the graduated elicitation methodology advocated by \citet{bowern2015} and others. The inclusion of verbal conjugation paradigms follows established practices in field linguistics for documenting inflectional morphology across typologically diverse languages \citep{payne1997}. Digital archiving and web-based dissemination of language documentation materials have become increasingly central to the field. Projects such as the Endangered Languages Documentation Programme (ELDP) at SOAS University of London and the Endangered Languages Archive (ELAR) have established standards for the digital preservation and open-access publication of endangered language corpora \citep{nathan2011}. Our work extends this tradition to a national scale, with the Multilingual Cloud platform providing web-based access to annotated audio and textual data for 42 language varieties.

\subsection{Computational Resources for Low-Resource Languages}

The severe imbalance in NLP resource availability across the world's languages has been quantitatively documented by \citet{joshi2020}, who proposed a six-class taxonomy ranging from Class 5 (resource-rich ``winners'' such as English) to Class 0 (``left-behind'' languages with virtually no digital presence). Under this taxonomy, the vast majority of Bangladesh's ethnic minority languages fall into Class 0 or Class 1, lacking even basic labelled datasets, let alone the pre-trained models and benchmarks available for higher-resource languages. This digital exclusion is compounded by the fact that many of these languages lack standardized orthographies, limiting the applicability of text-based NLP methods.

At the regional level, several large-scale corpus construction efforts have targeted South Asian languages, though these have overwhelmingly focused on the major national languages. The AI4Bharat IndicNLP Suite \citep{kakwani2020} provides monolingual corpora totalling 8.8 billion tokens across 11 Indian languages, along with pre-trained language models and evaluation benchmarks, but covers only major languages such as Hindi, Bengali, Tamil, and Telugu, omitting the smaller Tibeto-Burman, Austro-Asiatic, and Dravidian languages spoken in the region. Similarly, the Samanantar parallel corpus \citep{ramesh2022} and IndicTrans machine translation models have made significant advances for Indian language NLP but do not extend to extremely low-resource varieties.

A recent comprehensive survey by \citet{alam2025} on low-resource South Asian languages confirms that Tibeto-Burman and smaller Austro-Asiatic languages remain among the most severely underrepresented families in NLP research. The survey finds that while South Asian NLP has advanced substantially in recent years, the vast majority of datasets and models target a small set of high-resource Indic languages, with languages such as Manipuri, Mizo, and Bodo appearing only sporadically in available resources. \citet{liu2025} arrive at similar conclusions in their survey of NLP progress on Sino-Tibetan low-resource languages, finding that the Sino-Tibetan (including Tibeto-Burman) family as a whole has received minimal attention in the ACL Anthology despite comprising hundreds of languages.

Speech technology for low-resource languages faces even greater challenges. \citet{adams2019} demonstrate the feasibility of massively multilingual speech recognition but note that the performance gap widens dramatically for languages with less than 10 hours of transcribed audio. Transfer learning approaches for low-resource machine translation \citep{zoph2016,gu2018} and speech recognition have shown promise but require some amount of parallel or transcribed data as a starting point. Our 107-hour transcribed audio corpus across 40 languages provides precisely this foundational resource.

\subsection{The Gap Addressed by This Work}

Prior work on Bangladesh's linguistic minorities has been predominantly descriptive and community-specific, focusing on individual languages or small language groups (e.g., \citealp{akter2024}). No previous effort has attempted a systematic, parallel corpus construction across all four language families represented in Bangladesh at a national scale. The Multilingual Cloud corpus addresses this gap by providing for the first time a structured, annotated, publicly accessible dataset covering 42 language varieties with parallel elicitation, IPA transcription, and transcribed audio. This positions the corpus as a foundational resource for both computational linguistics research and community-driven language revitalization in one of the world's most linguistically underserved regions.

\section{Methodology}

\begin{figure}[!htb]
\centering
\includegraphics[width=\columnwidth,height=0.42\textheight,keepaspectratio]{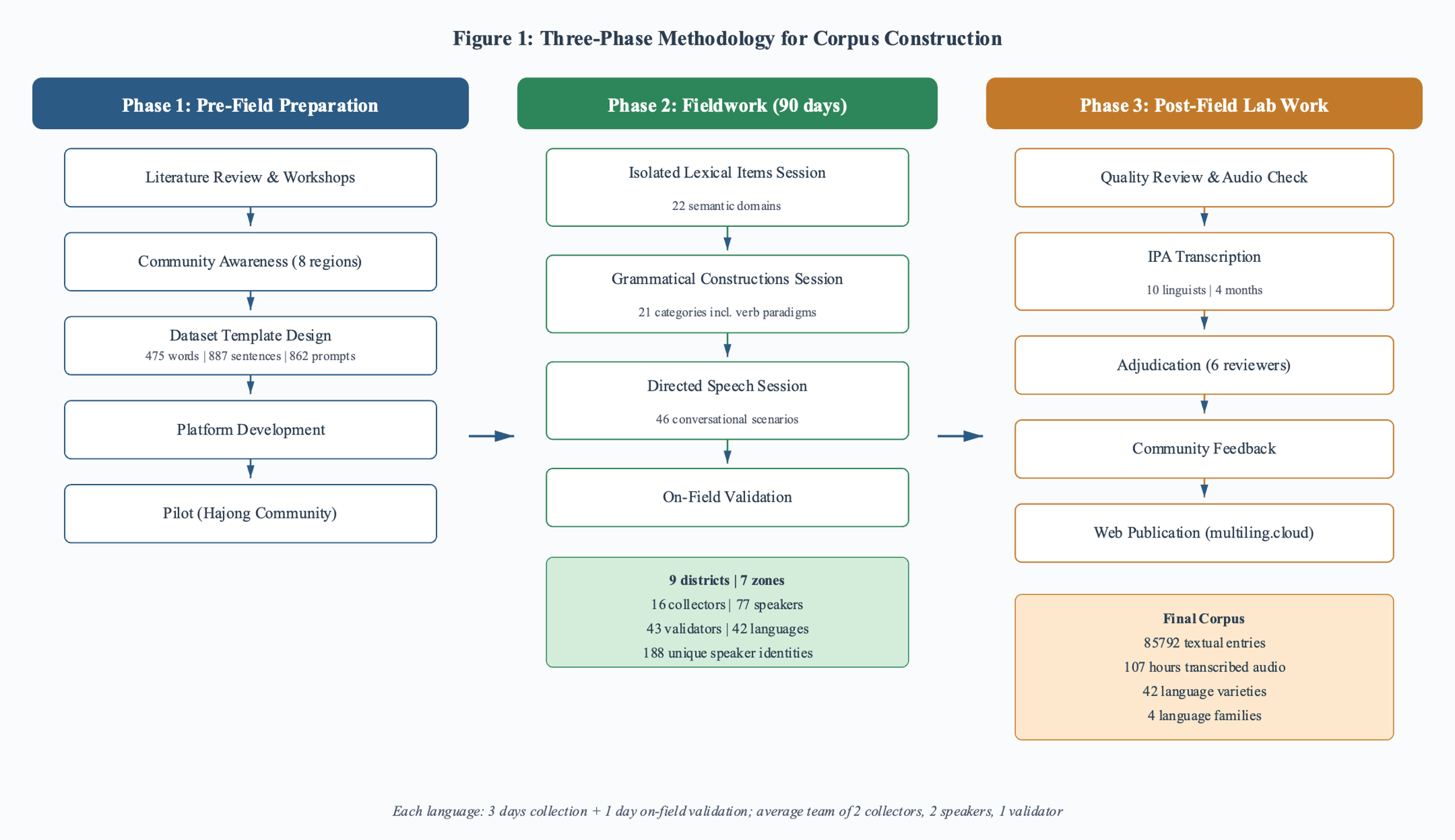}
\caption{Methodology for Corpus Construction}
\label{fig:methodology}
\end{figure}

The digital preservation of oral languages in this project followed three principal phases: Pre-Field Preparation, Fieldwork (Data Collection), and Post-Field Lab Work.

\subsection{Phase One: Pre-Field Preparation}

The pre-field preparation phase comprised five sequential activities designed to ensure methodological rigor and community engagement before data collection commenced. The project began with a comprehensive review of existing literature on language documentation methodologies, endangered language corpora, and the linguistic landscape of Bangladesh's ethnic minorities. Workshops were conducted to establish the scope of the data collection, define the target languages, and train the field team in elicitation techniques, audio recording protocols, and IPA transcription standards. This preparatory work informed the selection of 41 target languages across four language families (Tibeto-Burman, Austro-Asiatic, Dravidian, and Indo-European) and the identification of nine collection districts spanning seven zones of Bangladesh. After scope defining, community awareness meetings were held across eight regions of Bangladesh to introduce the project to minority language communities, explain its objectives, and receive feedback from community members and leaders. A predefined textual dataset was designed to ensure systematic and comparable data collection across all 41 languages. This template specified the exact words, sentences, and conversational topics to be elicited from speakers of each language, guaranteeing cross-linguistic comparability.

A custom fieldwork management platform was developed to support the data collection process. The platform featured two core capabilities: (i) an annotation interface that allowed data collectors to tag and categorize recorded speech data in real time, and (ii) an audio split function that enabled the segmentation of continuous recordings into individual entries aligned with the predefined elicitation items.

A one-day pilot fieldwork session was conducted with the Hajong community to test the complete data collection pipeline from elicitation and recording through annotation and validation. The pilot assessed the adequacy of the predefined template, the usability of the fieldwork management platform, the quality of audio recordings in field conditions, and the time required per language. Findings from the pilot were used to refine the elicitation protocol, adjust timing estimates for each language, and resolve technical issues before full-scale deployment.

\subsubsection{Predefined Dataset Template: Needs Analysis and Design}

The central challenge in building a parallel linguistic corpus across 41 typologically diverse languages is ensuring that the elicited material is both linguistically comprehensive and cross-linguistically comparable. The predefined dataset template was designed to address this challenge by specifying a standardized set of elicitation items at three levels of linguistic granularity: isolated lexical items (Words), grammatical constructions (Sentences), and extended discourse (Directed/Elicited Conversations). By eliciting the same set of items from every language, the template enables direct comparison of how different languages express the same concepts, grammatical relations, and communicative functions. This parallel structure is essential for typological analysis, machine translation training, and comparative linguistic research.

The word-level template specifies 475 unique lexical items distributed across 22 semantic domains. Each item was designed to be elicited as an isolated word or short phrase (1--2 words) in each target language. The domains, item counts, and representative examples are shown in Table~\ref{tab:words}.

\begin{table*}[t]
\centering
\small
\caption{The domains and examples of words}
\label{tab:words}
\begin{tabularx}{\textwidth}{lrX}
\toprule
\textbf{Domain} & \textbf{Items} & \textbf{Representative Examples} \\
\midrule
Verbal Words & 112 & to sit, to come, to go, to eat, to sleep, to run, to give, to take \\
Animals and Birds & 37 & cow, duck, tiger, bird, fish, snake, elephant, crow, goat \\
Words about Time & 34 & yesterday, today, tomorrow, now, then, later, morning, night \\
Kinship Terms & 32 & mother, father, grandfather, grandmother, uncle, aunt, son, daughter \\
Body Parts & 29 & hand, finger, mouth, eye, head, leg, ear, nose, teeth \\
Fruits & 27 & mango, jackfruit, banana, guava, coconut, papaya, litchi \\
Adjectives & 27 & good, beautiful, big, small, hot, cold, new, old, tall \\
Household Terms & 26 & home, shed, pole, door, roof, kitchen, bed, chair, pot \\
Words on Nature & 19 & village, water, river, mountain, forest, rain, sky, wind \\
Food Items & 17 & rice, lentil, salt, oil, fish, meat, milk, sugar, spice \\
Pronouns & 15 & this, that, here, there, who, what, I, you, he/she \\
Miscellaneous Words & 13 & family, discussion, question, answer, work, money, road \\
Periodic Words & 12 & time, morning, afternoon, evening, night, dawn, dusk \\
Person and Possessive Adjectives & 11 & myself, we, you, his, her, their, our, your \\
Emotive Words & 11 & smile, cry, joy, anger, fear, love, surprise, shame \\
Ordinal Numbers & 10 & first through tenth \\
Numerical Words & 10 & one through ten \\
Colors & 8 & white, black, red, blue, green, yellow, brown, grey \\
Livelihood Words & 8 & clothes, shoes, bangles, ornaments, tools, market \\
Name of 7 Days & 7 & Saturday through Friday \\
6 Seasons & 6 & summer, monsoon, autumn, late autumn, winter, spring \\
Interrogative Words & 5 & who, what, which, where, when \\
\bottomrule
\end{tabularx}
\end{table*}

The 112 verbal words constitute the largest single domain, reflecting the importance of documenting verb root forms before eliciting their conjugated paradigms in the Sentences component. The semantic domains were ordered to begin with more concrete, easily elicitable categories (animals, body parts, fruits) and progress toward more abstract domains (emotive words, miscellaneous), following established practices in field lexicography.

The sentence-level template specifies 887 unique sentence items across 21 categories, organized into two major subcategories: verbal conjugation paradigms and sentence-type elicitations. Five core verbs were selected for systematic conjugation across all person, number, tense, and aspect combinations relevant to the target languages. Each paradigm systematically varies person (1st, 2nd, 3rd), number (singular, dual, plural where grammatically relevant), and tense-aspect (present, past, future, habitual, progressive, perfective). For example, the ``saying'' paradigm elicits forms such as: ``I say,'' ``We two say,'' ``We say,'' ``You say,'' ``He/she says,'' ``I said,'' ``I will say,'' ``I am saying,'' and so on. The inclusion of dual number forms (e.g., ``we two say'') is deliberate, as many Tibeto-Burman and Austro-Asiatic languages in the dataset grammaticalize a dual-plural distinction that Bengali does not make.

Sixteen sentence types were included to document the full range of syntactic constructions, as shown in Table~\ref{tab:sentences}.

\begin{table*}[t]
\centering
\small
\caption{Pre-defined sentence types of the corpus}
\label{tab:sentences}
\begin{tabularx}{\textwidth}{lrX}
\toprule
\textbf{Sentence Type} & \textbf{Items} & \textbf{Design Purpose} \\
\midrule
Travel Manual & 47 & Formulaic questions and requests; documents interrogative and polite request strategies \\
Interrogative & 49 & Yes/no and wh-questions; documents question formation and word order \\
Assertive-Negative & 46 & Negative declaratives; documents negation placement and morphology \\
Sensory & 48 & Perception and sensation descriptions; documents sensory verb constructions \\
Assertive-Affirmative & 33 & Positive declaratives; baseline clause structure documentation \\
Case Markers & 38 & Noun phrases with explicit case relations; documents case marking systems \\
Imperative & 29 & Commands and requests; documents imperative mood morphology \\
Compound & 24 & Coordinated clauses; documents conjunction strategies \\
Complex & 22 & Subordinated clauses; documents relativization and complementation \\
Simple & 18 & Basic clause structures; documents core argument ordering \\
Speculative & 15 & Epistemic modality; documents evidential and modal marking \\
Name of 12 Months & 12 & Calendrical terms; documents indigenous temporal systems \\
Negative-Interrogative & 12 & Negative questions; documents interaction of negation and interrogation \\
Exclamatory & 11 & Emotional expressions; documents exclamative constructions \\
Optative & 7 & Wishes and blessings; documents optative/subjunctive mood \\
Complex-Compound & 4 & Multi-clause constructions; documents maximum syntactic complexity \\
\bottomrule
\end{tabularx}
\end{table*}

The sentence types are ordered roughly by frequency in the dataset, but were designed to progress from simpler to more complex constructions during elicitation sessions, following the principle that speakers produce more natural data when complexity is gradually increased.

The directed speech component specifies 46 active conversational scenarios (with 8 additional topics later excluded from the final dataset), totaling 862 unique prompts. Each scenario provides speakers with a situational context and a sequence of dialogue turns or narrative prompts, which the speaker renders in their native language.

The scenarios were organized into eight thematic domains:

\textbf{Domestic and Social Life:} Conversation about marriage (37 prompts); teaching a daughter to cook (31); the old banyan tree of the village (33); meeting someone new (25); exchanging household goods (24); family inquiries (18); mother's instructions about grocery shopping (26); conversation at the grocery store (16); eating at a restaurant (20); conversation between father and daughter about going to a fair (16); and father-son conversation about repairing the house (26).

\textbf{Rural Economy and Livelihood:} Conversation on agriculture (26 prompts); husbandry of domestic animals (27); conversation about lost cattle (24); drinkable water scarcity (22); and village market (18).

\textbf{Community and Culture:} Organising a festival in the area (28 prompts); festival time, now and then (23); and use of mobile phones (10).

\textbf{Education and Public Services:} Going to school for the first time (24 prompts); conversation with a teacher about enrolling a son in school (23); conversation at the bus counter (16); assistance at the police station (17); and assisting an injured person (17).

\textbf{Health:} Treatment of children with diarrhoea (19 prompts); health advice for a newborn baby (12); and taking care of a pregnant mother (10).

\textbf{Folk Stories:} The story of the wind and the sun (18 prompts); the story of the ant and the dove (12); the story of honey and the ant (10); the story of the tiger and the pig (8); and the story of the crow's thirst (8).

\textbf{Descriptive Passages and Personal Narratives} (4 scenarios; ${\sim}$2085 entries): Chirping of birds (25 prompts); childhood (18); childhood garden (11); and Nina's letter (10).

\textbf{Thematic Essays:} Father (21 prompts); mother (12); forest (15); water (13); agriculture (17); family (15); language (15); fish (9); and Bangladesh (11).

The conversational scenarios were selected based on three criteria: (i) cultural relevance to the daily lives of minority communities in rural and semi-urban Bangladesh; (ii) diversity of communicative functions---the scenarios collectively require speakers to request, narrate, instruct, persuade, describe, express emotion, and perform social rituals; and (iii) elicitation of domain-specific vocabulary that may not emerge from the isolated word or sentence components (e.g., agricultural terminology, medical vocabulary, kinship address forms in context).

Eight additional scenarios were piloted but subsequently excluded from the final dataset due to insufficient coverage across languages: conversation on seed preservation (60 entries), father-son talk about a bicycle (28), sports---now and then (26), visiting uncle's house (24), neighbor's hospital treatment (22), teaching math class (20), finding a community clinic (20), and the foolish deer story (10). These 210 entries remain in the dataset with exclusion markers for transparency.

\begin{table}[t]
\centering
\small
\caption{Summary of the Predefined Template}
\label{tab:summary}
\begin{tabular}{lrrr}
\toprule
\textbf{Category} & \textbf{Topics} & \textbf{Unique} & \textbf{Total} \\
 & & \textbf{Items} & \textbf{Entries} \\
\midrule
Isolated Lexical & 22 & 475 & 18344 \\
Items (Words) & & & \\
Grammatical Constr. & 21 & 887 & 34204 \\
(Sentences) & & & \\
Directed Speech & 46 (+8) & 862 & 33244 \\
\midrule
\textbf{Total} & \textbf{97} & \textbf{2224} & \textbf{85792} \\
\bottomrule
\end{tabular}
\end{table}

The template was designed to yield approximately 2000 entries per language. In practice, per-language totals range from 1217 (Soura) to 4145 (Telugu), with languages that include multiple dialectal varieties (Marma, Kheyang, Telugu, Kokborok/Usui) exceeding the target due to data from additional speakers representing different varieties.

\subsection{Phase Two: Fieldwork (Data Collection)}

The fieldwork was conducted over 90 days across seven zones of Bangladesh, covering nine districts: Moulvibazar, Bandarban, Rangamati, Rajshahi, Mymensingh, Dhaka, Dinajpur, Netrokona, and Cox's Bazar. Districts were selected to match the geographic distribution of minority language communities, with the Chittagong Hill Tracts (Bandarban, Rangamati, Cox's Bazar) and the Sylhet division (Moulvibazar) contributing the largest shares of the data.

A total of 16 data collectors gathered speech data from 77 speakers across all 41 languages. Each language was allocated 3 days for primary data collection and 1 day for on-field validation. On average, each language required 3.3 days with a team of 2 data collectors working with 2 speakers and 1 validator. Individual collectors were assigned to work across multiple languages (up to 17 languages per collector) to ensure cross-language methodological consistency.

For each language, the elicitation followed the three-part structure of the predefined template:

\begin{enumerate}[leftmargin=*,topsep=2pt,itemsep=2pt]
\item \textbf{Words session.} Speakers were presented with Bengali stimulus words (organized by semantic domain) and asked to produce the equivalent in their language. Each word was recorded individually, with the data collector annotating the entry using the fieldwork management platform.

\item \textbf{Sentences session.} Speakers were presented with Bengali sentences and asked to produce the equivalent in their language. Verbal conjugation paradigms were elicited systematically, with the collector stepping through each person-number-tense combination. Sentence-type items were presented in order of increasing complexity.

\item \textbf{Directed speech session.} Speakers were given situational prompts for each conversational scenario and asked to produce naturalistic dialogue or narrative. For dialogue-based topics, two speakers sometimes alternated roles; for narrative and essay topics, a single speaker produced the full text.
\end{enumerate}

\subsubsection{On-Field Validation}

On-field validation was conducted by a different speaker from the same language community, distinct from the primary data speakers. The validator listened to the recorded data and verified the accuracy, naturalness, and representativeness of the language produced. This step ensured that the collected data authentically represented the target language and was not unduly influenced by individual speaker idiosyncrasies or code-switching with Bengali.

\subsubsection{Speaker Demographics}

A total of 77 speakers and 43 validators participated in the fieldwork data collection, with 188 unique speaker identities represented in the final textual corpus. The number of speakers per language ranges from 1 to 10, with a median of 4. Marma has the highest speaker count (10), followed by Koda (9) and Kheyang (8). One language (Gurkhali) is represented by a single speaker.

\subsection{Phase Three: Post-Field Lab Work}

All audio recordings were systematically reviewed for recording quality, completeness, and adherence to the elicitation protocol. Entries with audio quality issues (background noise, clipping, unintelligible speech) were flagged for re-recording or exclusion. The completeness check verified that all predefined template items had been elicited for each language.

Phonetic transcription of all collected speech data was carried out by a team of 10 linguists over a period of 4 months. Each linguist was assigned specific languages based on their regional and linguistic expertise. The transcription followed IPA conventions, with language-specific diacritics and symbols used where standard IPA categories were insufficient.

An additional 6 adjudicators independently reviewed the IPA transcriptions for consistency and accuracy. The adjudication process involved spot-checking a sample of transcriptions from each language, resolving discrepancies between the original transcription and the adjudicator's assessment, and establishing language-specific transcription conventions where ambiguity existed. This two-tier process---independent transcription followed by adjudication---ensured a high level of phonetic annotation quality across the 41 languages.

The finalized dataset for each language was presented to members of the respective language community for review and acceptance. Community feedback was incorporated where it identified errors in transcription, translation, or cultural appropriateness of the elicited content.

The completed dataset was published through the Multiling.cloud web platform, making it publicly accessible for research and community use. The platform provides searchable access to the textual data (Bengali, English, IPA) alongside the segmented audio recordings for each entry.

\section{Dataset Description}

This paper presents a large-scale, multimodal linguistic dataset covering 42 minority and indigenous languages spoken in Bangladesh. The dataset was constructed through a systematic field data collection effort across nine districts in seven zones of Bangladesh, involving 16 data collectors, 77 speakers, and 43 validators over a 90-day fieldwork period. The resulting corpus consists of two complementary components: (i) a structured textual dataset of 85,792 annotated entries, each containing a Bengali stimulus text, its English translation, and a phonetic transcription in the International Phonetic Alphabet (IPA); and (ii) approximately 107 hours of transcribed audio recordings across 40 languages. The dataset is designed to serve as a foundational resource for computational linguistics, language documentation, typological research, and the development of natural language processing (NLP) and speech technology tools for extremely low-resource languages.

Each language contains a median of 1961 textual entries, with a minimum of 1217 (Soura) and a maximum of 3911 (Kheyang). Languages with multiple dialectal or regional varieties---Kheyang (3911), Marma (3903), and Kol (2443)---exhibit larger entry counts due to data collected from multiple dialect groups or regions (see Section~\ref{fig:methodology}).

Note on language classification: Telugu and Madrasi (Telugu) are maintained as separate entries in the dataset. Both communities speak the Telugu language, but they represent distinct sociolinguistic populations. The Telugu community in Moulvibazar (2162 entries, 6 speakers) and the Madrasi community in Dhaka (1983 entries, 4 speakers) are kept separate to preserve these sociolectal distinctions for future comparative research. Kheyang encompasses the Laitu and Kontu dialect varieties. Usui, while closely related to Kokborok as a dialect of the Tripura language complex, is maintained as a separate language entry in the textual corpus; their combined audio recording is noted in the audio corpus.

\subsection{Data Structure}

Each entry in the textual corpus consists of ten fields, as shown in Table~\ref{tab:structure}.

\begin{table}[t]
\centering
\small
\caption{Data Structure}
\label{tab:structure}
\begin{tabularx}{\columnwidth}{lX}
\toprule
\textbf{Field} & \textbf{Description} \\
\midrule
Language Name & Target minority language in English \\
Serial No & Sequential identifier within each language subset \\
Data Type & Collection category: Words, Sentences, or Directed/Elicited \\
District & Geographic location where the data was recorded \\
Data Collector & Name of the trained field researcher \\
Speaker & Name of the native language consultant \\
Topic & Thematic category of the elicitation prompt in English \\
Text in Bangla & Source stimulus or target-language rendering in Bengali script \\
Text in English & Parallel English translation \\
IPA Transcription & Phonetic rendering in the International Phonetic Alphabet \\
\bottomrule
\end{tabularx}
\end{table}

\subsection{Per-Language Data Coverage}

\begin{figure}[!htb]
\centering
\includegraphics[width=\columnwidth,height=0.45\textheight,keepaspectratio]{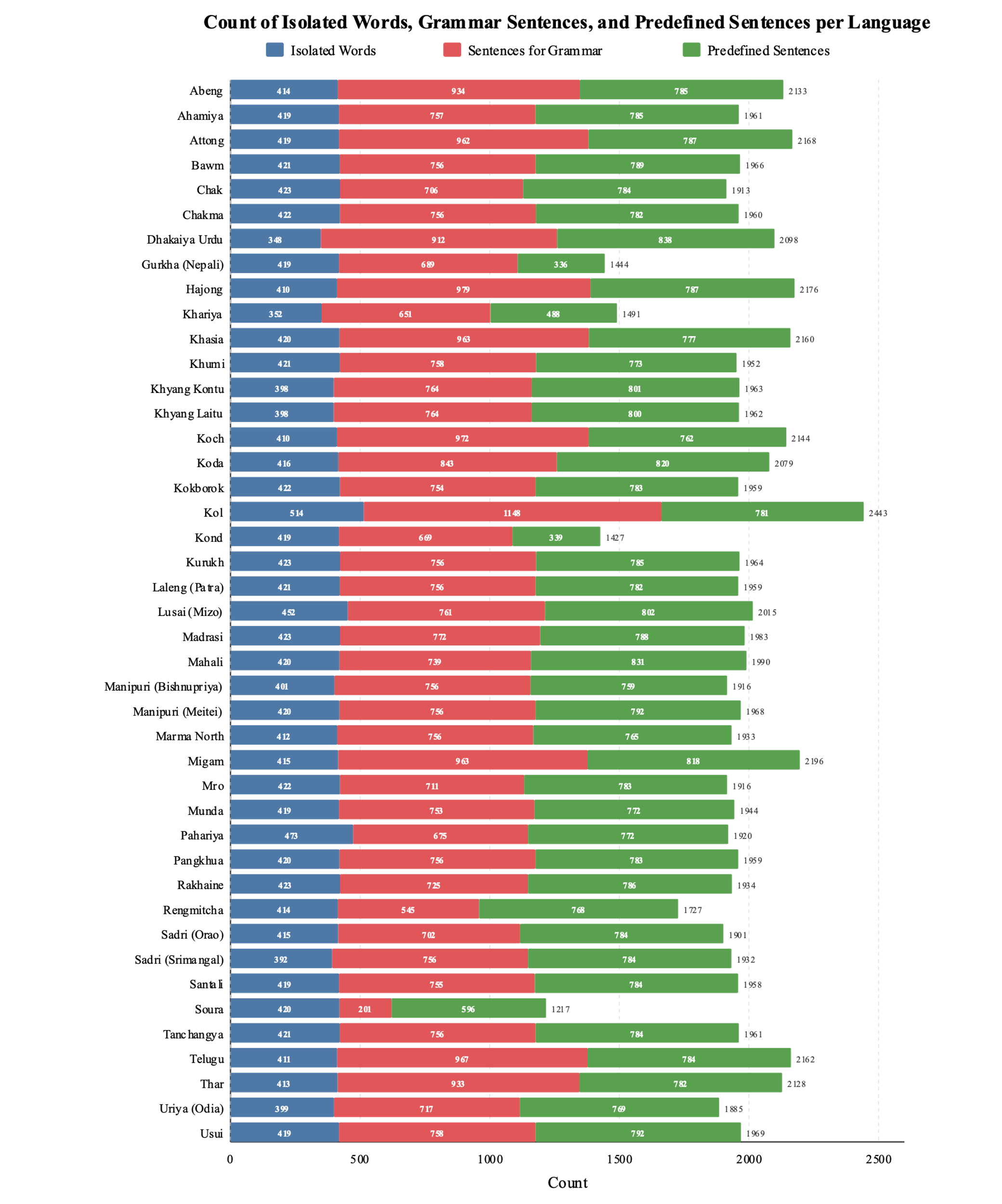}
\caption{Per-Language Data Coverage}
\label{fig:coverage}
\end{figure}

The predefined elicitation template specified 475 unique word items across 22 semantic domains, 887 unique sentence items across 21 grammatical categories, and 862 unique prompts across 46 active conversational scenarios. Word-level collection shows the highest consistency. Twenty-nine out of 42 languages (69.0\%) achieved full 22/22 topic coverage. The remaining 13 cover 19--21 topics, with Khariya (19/22) and Dhakaiya Urdu (20/22) being the lowest. The median word count per language is 420, closely matching the 475-item template. Multi-dialect languages---Marma (824), Khyang (796), and Kol (514)---show elevated counts reflecting data from multiple dialect groups. Sentence-level coverage is more variable. Only 6 languages achieved full 21/21 topic coverage (Garo Attong, Garo Migam, Hajong, Khasia, Koch, and Telugu). Twenty-six languages cover 19/21, indicating that Verbal Conjugation paradigms for Hearing and Bringing were not uniformly collected. Soura shows the largest gap at 10/21 topics (201 entries), followed by Kond (12/21) and Gurkha (15/21). Twenty-seven languages (64.3\%) achieved full 46/46 active topic coverage. Most others cover 44--45 topics. Three languages show substantially reduced directed data: Gurkha (20/46 topics, 336 entries), Kond (17/46, 339 entries), and Khariya (28/46, 488 entries), attributable to shorter fieldwork duration. Two languages significantly exceed the corpus-wide median due to dialectal varieties: Khyang (3911) and Marma (3903). Four fall notably below the target: Soura (1217), Kond (1427), Gurkha (1444), and Khariya (1491), with reduced entries accompanied by lower topic coverage in Sentences and Directed categories.

\subsection{Geographic Coverage}

Data were collected across nine districts spanning seven zones of Bangladesh. Moulvibazar in the Sylhet Division contributed the largest share with 18,117 entries across 10 languages and 44 speakers, reflecting the concentration of tea garden worker communities in this region. The Chittagong Hill Tracts region, comprising Bandarban (17,378 entries, 9 languages, 34 speakers), Rangamati (15,159 entries, 8 languages, 33 speakers), and Cox's Bazar (1913 entries, 1 language, 3 speakers), collectively accounts for 34,450 entries, reflecting the high linguistic diversity of this area. Rajshahi contributed 10,333 entries across 5 languages and 29 speakers from northwestern Bangladesh. Mymensingh (6473 entries, 3 languages, 14 speakers) and Dhaka (6209 entries, 3 languages, 12 speakers) represent north-central and central Bangladesh respectively. Dinajpur contributed 5866 entries across 3 languages and 12 speakers, while Netrokona provided 4344 entries across 2 languages and 7 speakers. Two languages were collected from two districts: Marma (Rangamati and Bandarban, reflecting its northern and southern dialectal varieties) and Kheyang (Bandarban and Rangamati, reflecting the Laitu and Kontu dialects).

\subsection{Audio Corpus}

\begin{figure}[!htb]
\centering
\includegraphics[width=\columnwidth,height=0.45\textheight,keepaspectratio]{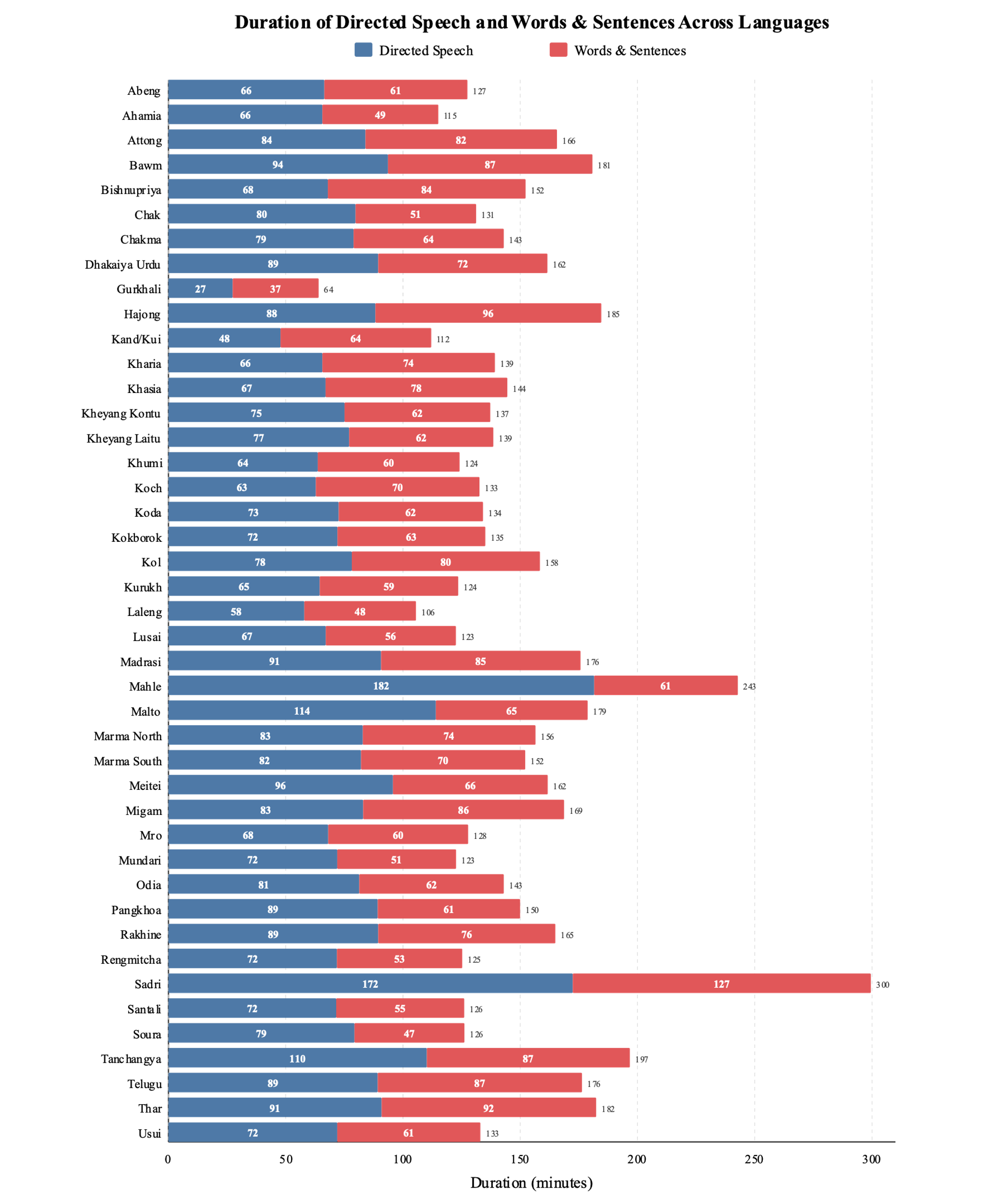}
\caption{Audio Corpus Duration by Language}
\label{fig:audio}
\end{figure}

The audio component comprises approximately 107 hours of transcribed recordings across 40 languages. Kokborok and Usui share a combined audio recording, as do Telugu and Madrasi (Telugu).

\subsection{Dialectal and Regional Varieties}

Several languages in this dataset include data from multiple dialectal or regional varieties, reflecting the sociolinguistic reality that these varieties are mutually intelligible and recognized as dialects of a common language by their speaker communities. Marma includes two regional varieties: a northern variety collected from Rangamati district and a southern variety collected from Bandarban district, yielding a combined total of 3903 entries across 10 speakers. Kheyang includes the Laitu and Kontu dialect varieties, combined into a single dataset of 3911 entries collected from 8 speakers across the Bandarban and Rangamati districts. Telugu and Madrasi (Telugu) are linguistically the same language but are maintained as separate entries to preserve sociolinguistic distinctions. Telugu represents the community in Moulvibazar, while Madrasi (Telugu) represents the community in Dhaka. Their audio recordings are listed as a combined entry in the audio corpus. Kokborok and Usui are closely related varieties of the Tripura language complex. While their audio recordings were captured together, they are maintained as separate entries in the textual corpus to preserve dialectal distinctions.

Sadri is represented as two distinct community varieties: Sadri (Oraon) from North Bengal and Sadri (Srimangal) from the tea worker communities of the Sylhet division. Garo is represented by three dialect varieties: Abeng, Attong, and Migam. Manipuri is represented by two distinct varieties: Meitei and Bishnupriya.

\subsection{Data Quality}

The textual corpus demonstrates a high level of completeness. Bengali text is present for 100\% of all 85,792 entries. English translations are missing for only 210 entries (0.24\%), all within the Directed/Elicited category. IPA transcriptions are absent for 860 entries (1.00\%), distributed as: 384 in Directed/Elicited, 352 in Sentences, and 124 in Words. Missing IPA entries are concentrated in Mundari (404 entries, 47.0\% of all IPA gaps). All other languages show IPA completion rates above 95\%. The IPA transcription process involved 10 linguists working over 4 months, with 6 adjudicators conducting independent quality review.

\FloatBarrier
\section{Discussion}

This section interprets the principal findings of the Multilingual Cloud Corpus in relation to the broader challenges of endangered language documentation, computational resource development, and digital preservation in linguistically diverse developing countries.

\subsection{Endangered Language Documentation: Patterns and Implications}

The documentation of 14 endangered languages constitutes one of the most significant outcomes of this work, and the corpus data reveal several patterns with implications for language endangerment theory and revitalization practice.

The per-language audio distribution shows a predictable correlation between community vitality and data yield: languages with larger, more geographically concentrated speaker populations (Marma, 309 minutes; Sadri, 300 minutes; Kheyang, 274 minutes) produced substantially more audio material than critically endangered languages with dispersed or elderly speakers (Gurkhali, 64 minutes; Laleng, 106 minutes; Kand/Kui, 112 minutes). Yet critically, even the most endangered languages in the corpus yielded sufficient material for meaningful linguistic analysis. Rengmitcha, with approximately six speakers all over age 60 near Alikadam in the southern Chittagong Hill Tracts, produced 1727 textual entries and 125 minutes of audio across all three elicitation tiers. This outcome demonstrates that structured elicitation protocols can extract substantial documentation even from languages at the very edge of extinction, provided that fieldwork is planned with sensitivity to speaker capacity and community dynamics.

The case of Rengmitcha warrants particular attention as a case study in terminal language documentation. Classified as a Khomic language of the Kuki-Chin branch within the Tibeto-Burman family \citep{peterson2017}, Rengmitcha has lost intergenerational transmission entirely: younger family members know at most a few words, and the remaining speakers are elderly and geographically isolated. The corpus data for Rengmitcha, while necessarily limited in speaker diversity, represent what may be the final systematic recording of this language in any form. Similarly, Kharia, spoken by as few as five individuals in the tea garden communities of Sreemangal, Moulvibazar, yielded 1491 entries and 139 minutes of audio. These recordings constitute irreplaceable documentation of languages whose phonological systems, grammatical structures, and lexical inventories would otherwise be entirely lost to scholarship and to the communities themselves.

A counterintuitive finding merits discussion: several languages with established writing systems, including Chak, Koda, Kol, and Pangkhua, are nonetheless classified as endangered by \citet{imli2018}. This observation challenges the common assumption that literacy ensures language vitality. The data suggest that in the absence of active intergenerational transmission, institutional support, and domains of use beyond ceremonial or literary contexts, written traditions alone are insufficient to sustain a language. This finding aligns with broader sociolinguistic research on the decoupling of literacy from vitality \citep{crystal2000,harrison2007} and underscores the importance of digital documentation as a complementary preservation strategy even for languages with written forms.

\subsection{Implications for Computational Linguistics and NLP}

The Multilingual Cloud Corpus is positioned to address a critical resource gap for NLP in extremely low-resource settings. The 107-hour transcribed audio component across 40 languages is significant in light of \citeauthor{adams2019}'s~(\citeyear{adams2019}) finding that ASR performance degrades sharply for languages with fewer than 10 hours of labelled data: while per-language audio duration in our corpus ranges from 64 minutes (Gurkhali) to 352 minutes (Telugu/Madrasi), even the lower end of this range provides sufficient seed data for transfer learning and few-shot approaches that have shown promise in recent low-resource speech research \citep{zoph2016,gu2018}.

The parallel structure of the textual corpus, with Bengali-English-IPA alignment across all entries, is directly amenable to several computational tasks. The word-level component (18,344 entries across 22 semantic domains) can serve as training data for bilingual lexicon induction and cross-lingual word embedding alignment. The sentence-level component (34,204 entries with systematic verbal paradigms) supports morphological analysis and paradigm learning. The directed speech component (33,244 entries across 46 scenarios) supports conversational NLP and discourse modelling. These applications position the corpus as a foundational resource that can catalyse the development of downstream NLP tools, including morphological analyzers, machine translation systems, and speech technology, for languages that \citet{joshi2020} classify as computationally ``left behind.''

The dataset's value is further enhanced by its parallel structure across four genetically unrelated language families. Cross-family transfer learning, in which models trained on better-resourced languages within a family are adapted to lower-resource relatives, represents a promising but underexplored direction for the languages in this corpus. The Tibeto-Burman sub-corpus alone, covering 21 varieties across six branches, constitutes a resource of unusual density for this family, which \citet{liu2025} identify as one of the most neglected in NLP research.

\subsection{Digital Preservation and the Multilingual Cloud Platform}

The resulting dataset is publicly accessible via the Multilingual Cloud platform (\url{https://multiling.cloud}), launched in July 2025 under the EBLICT project of the Bangladesh Computer Council. The platform provides an interactive language map, language-by-language browsing with descriptive profiles, audio playback aligned with IPA and bilingual translations, and topic-based navigation across 7177 thematic categories. It represents the first national-scale digital archive of ethnic languages in Bangladesh in audio-visual form.

The significance of the platform extends beyond data hosting. For many of the documented communities, the Multilingual Cloud represents the first time their language has been rendered in a structured digital form accessible to speakers outside the immediate community. The accompanying digital fonts and universal keyboard, supporting all scripts used by the documented languages, address a fundamental barrier to digital participation: the inability to write one's own language on a computer or phone. In this respect, the platform functions not only as a research archive but as an enabling infrastructure for digital literacy and community-driven language use in domains (education, social media, messaging) where these languages have been historically absent.

Bangladesh's endangered languages face a dual challenge that the platform is designed to address: declining intergenerational transmission threatens their survival as living languages, while the absence of digital resources threatens their survival even as documentary records. The multiling.cloud platform mitigates the second threat by providing durable, publicly accessible documentation that preserves vocabulary, grammar, phonology, and connected speech for future generations, regardless of whether the languages continue to be spoken. This approach, combining scientific documentation with open-access digital infrastructure, offers a model that other linguistically diverse nations in South and Southeast Asia could adapt to their own contexts.

\FloatBarrier
\section{Limitations}

This study has several limitations that should be acknowledged:

\begin{enumerate}[leftmargin=*,topsep=2pt,itemsep=4pt]
\item \textbf{Uneven per-language coverage:} The amount of data collected varies across languages due to differences in speaker availability, geographic accessibility, and community engagement. Languages with very few remaining speakers (e.g., Rengmitcha, Kharia) necessarily have less data than more widely spoken languages. Four languages fall below the target threshold: Soura (1217), Kand/Kui (1427), Gurkhali (1444), and Kharia (1491).

\item \textbf{Speaker representation:} With an average of approximately 2 speakers per language in the initial fieldwork phase, individual speaker variation (age, gender, dialectal background) may not be fully captured for all languages. One language (Gurkhali) has only a single consultant.

\item \textbf{IPA transcription reliability:} While transcriptions were performed by trained phoneticians, systematic inter-annotator agreement metrics were not computed. Some phonemic distinctions may be inconsistently represented, particularly for languages lacking prior phonological descriptions. The IPA transcriptions were produced by linguists working with native speakers rather than by trained phoneticians embedded in each community.

\item \textbf{Completeness of phonological documentation:} While the dataset captures 97 topics across three categories, it may not fully represent the complete phonological, morphological, and syntactic inventory of each language.
\end{enumerate}

\FloatBarrier
\section{Conclusion and Future Work}

We have presented a comprehensive account of the digitization of oral and low-resource languages of Bangladesh, from initial scoping through public release on the Multilingual Cloud platform. Future work will focus on several directions. First, expanding the corpus with additional speakers, particularly for under-represented languages, to increase dialect coverage and speaker diversity. Second, developing computational tools, including automatic speech recognition, text-to-speech synthesis, and machine translation models, using the Multilingual Cloud corpus as training data. Third, building language learning applications and interactive tools to support community-driven revitalization efforts. Fourth, extending the documentation to capture additional cultural and linguistic features such as oral narratives, ceremonial language, and discourse-level phenomena. Finally, the platform will be maintained and expanded to preserve the intangible cultural heritage embedded in these languages for future generations. The Multilingual Cloud platform stands as evidence that large-scale, government-supported digital language documentation is feasible in a developing country context, and offers a replicable model for other linguistically diverse nations in South and Southeast Asia.

\section*{Acknowledgments}

This work was conducted under the Enhancement of Bangla Language in ICT through Research and Development (EBLICT) project, implemented by the Bangladesh Computer Council (BCC) under the ICT Division, Ministry of Posts, Telecommunications, and Information Technology, Government of the People's Republic of Bangladesh. We gratefully acknowledge the financial support of the Government of Bangladesh. We thank the 214 native speakers who contributed their linguistic knowledge, the 43 community validators, the 16 field data collectors, and the ethnic language experts and linguistics graduates from the University of Dhaka who participated in fieldwork. We also thank Dream 71 Bangladesh Limited and other technical partners for software development support. We are deeply grateful to the ethnic communities across Bangladesh who welcomed research teams into their homes and shared their languages and cultural knowledge.

The following individuals contributed to the design, fieldwork, data processing, and technical implementation of the Multilingual Cloud Corpus. Roles reflect the primary responsibilities held during the project.

\begin{table*}[t]
\centering
\small
\caption{Project Contributors}
\label{tab:contributors}
\begin{tabularx}{\textwidth}{llX}
\toprule
\textbf{Name} & \textbf{Designation} & \textbf{Role} \\
\midrule
Michael Mridul Kanti Sangma & Ethnic Language Expert & Lead contributor for gathering linguistic diversity information from the field; data collection; language profiling; IPA standard definition and transcription \\
Charu Haq & Ethnic Language Expert & Lead contributor for compiling linguistic diversity information from the field; compiling language profiles; data collection \\
Sharmin Akter Shova & Lead Linguist & IPA standard definition and transcription supervision; data collection \\
Samar Michael Soren & Ethnic Language Specialist & Field data collection; language profiling; IPA standard definition and transcription \\
Ribeng Dewan & Data Collection Specialist & Lead contributor for gathering linguistic diversity information from the field; data collection; language profiling \\
Eduard Soren & Associate Linguist & Lead contributor for gathering linguistic diversity information from the field; data collection; language profiling; IPA standard definition and transcription \\
Khairun Nahar Insita & Linguist & Dataset preparation; language profiling; IPA standard definition and transcription; data collection \\
Sajia Morshed Mehnaz & Linguist & Lead contributor for dataset preparation \\
Khairul Islam Kiron & Associate Linguist & Data collection; language profiling; IPA standard definition and transcription \\
Tanzina Mousumi Chowdhury & Associate Linguist & Data collection; IPA transcription \\
Nusrat Jahan Propa & Associate Linguist & Data collection; IPA transcription \\
Umme Salma Sharia & Associate Linguist & Data collection; IPA transcription \\
Nusrat Jahan Mumu & Associate Linguist & Data collection; IPA transcription \\
Sajib Barman & Associate Linguist & Data collection; IPA transcription \\
Nobonita Lodh Hridi & Documentation Associate & Data collection; writing technical documents \\
Champa Naidu & Field Associate & Data collection \\
Muhammad Shahidul Islam & Documentation Officer & Page layout and formatting of materials \\
Foysal Rafat & Documentation Associate & Videography; web design \\
Abid Sarkar Sohag & Video Production Producer & Video production \\
Nazmul Gani & Project Manager & Coordination and management of fieldwork \\
Chowdhury Md. Mesbahul Ibne Munir & Program Officer & Logistics and communication management \\
Helal Uddin Hejazi & Technology Mgmt. Specialist & Core project executor; design and quality assurance of fieldwork software \\
Mohammad Mamun Or Rashid & Consultant, EBLICT Project & Project planner; chief research and scientific decision-maker; quality assurance on behalf of EBLICT \\
\bottomrule
\end{tabularx}
\end{table*}

\section*{Ethical Considerations and Consent}

Language documentation inherently involves working with marginalized communities and culturally sensitive materials. The following ethical practices were observed:

\textbf{Community consent:} Community awareness meetings were held prior to data collection at all sites. The objectives, methods, and intended use of the data were explained to community leaders and participants. Informed consent was obtained from all speakers and validators.

\textbf{Community participation:} Speakers and validators were recruited from within their own communities, ensuring that documentation was conducted with community involvement rather than as an extractive process. In many cases, researchers embedded within communities during fieldwork.

\textbf{Data sovereignty:} The resulting data is hosted on a publicly accessible government platform (multiling.cloud) with the stated aim of preserving and promoting the languages for the benefit of the documented communities and the public.

\textbf{Cultural sensitivity:} The thematic design of the dataset was developed to be culturally appropriate, covering topics relevant to daily life, traditions, and community practices as identified in consultation with community members.

\textbf{Intellectual property and publication:} This paper has been published with the prior necessary permission of the Project Director of the EBLICT project, who retains all intellectual property and distribution rights for this publication on behalf of the Bangladesh Computer Council.

\textbf{Ongoing considerations:} Questions of long-term data governance, community control over linguistic resources, and the right to withdraw data remain important and should be addressed in future phases of the project.

\section*{Data Availability}

The complete dataset is publicly available at \url{https://multiling.cloud}. The platform provides open access to audio recordings, IPA transcriptions, Bangla and English translations, and linguistic metadata for all 41 documented language varieties. There are no access restrictions. Researchers seeking to use the data for computational purposes are encouraged to contact the EBLICT project at the Bangladesh Computer Council (\texttt{pdeblict@bcc.net.bd}) for additional information regarding data formats and bulk access.

\bibliographystyle{plainnat}
\bibliography{references}

\end{document}